# Model-Based Sensor Diagnostics for Robotic Manipulators

Astha Kukreja

*Abstract*—Ensuring the safe and reliable operation of collaborative robots demands robust sensor diagnostics. This paper introduces a methodology for formulating model-based constraints tailored for sensor diagnostics, featuring analytical relationships extending across mechanical and electrical domains. While applicable to various robotic systems, the study specifically centers on a robotic joint employing a series elastic actuator. Three distinct constraints are imposed on the series elastic actuator: the Torsional Spring Constraint, Joint Dynamics Constraint, and Electrical Motor Constraint. Through a simulation example, we demonstrate the efficacy of the proposed model-based sensor diagnostics methodology. The study addresses two distinct types of sensor faults that may arise in the torque sensor of a robot joint, and delves into their respective detection methods. This insightful sensor diagnostic methodology is customizable and applicable across various components of robots, offering fault diagnostic and isolation capabilities. This research contributes valuable insights aimed at enhancing the diagnostic capabilities essential for the optimal performance of robotic manipulators in collaborative environments.

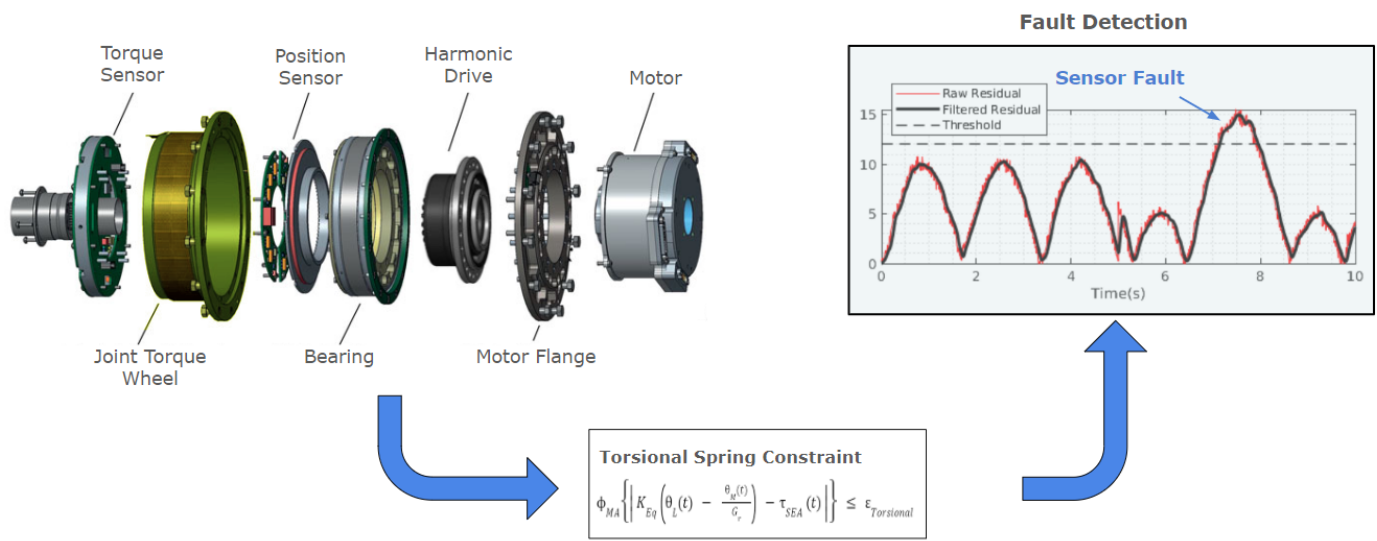

## I. INTRODUCTION

Robotics manipulators are widely used in automation for various industrial applications, including manufacturing, assembly, and packaging tasks. As industrial tasks become increasingly automated, the role of robotics manipulators in industrial settings continues to expand. New paradigms in industrial robotics no longer require physical separation between robotic manipulators and humans, fostering a collaborative environment where humans and robots work together to optimize production. Robotic manipulators are sophisticated actuators that rely on feedback from various sensors to function effectively. These sensors, such as encoders, position sensors, torque sensors, and temperature sensors, provide crucial information about the manipulator's state and environment. Sensor failure can have severe consequences, leading to downtime, loss of production, and potentially hazardous situations, especially in human-robot collaborative environments.

Sensor diagnostics plays a critical role in ensuring the reliability and safety of robotic manipulators. It involves the detection and identification of sensor faults to initiate appropriate fail-safe procedures, bringing the robot to a safe stop or safely shutting it down. Sensor diagnostics is essential for preventing unpredictable robot motions that could endanger humans working alongside the robot. [1] highlight the significance of sensor diagnostics in industrial applications, emphasizing the need for robust and reliable sensor systems. The safety concerns associated with human-robot collaboration and the need for effective sensor diagnostics to prevent hazardous situations is emphasized in [2].

Various methodologies exist for diagnosing sensor faults, including the straightforward identification of out-of-range sensor values. Additionally, the implementation of redundant sensing mechanisms can be employed to cross-verify values. [3] proposed a sensor value validation approach based on systematic exploration of sensor redundancy based on causal relations and their interrelations within sensor redundancy graphs. An introduction to sensor signal validation in redundant measurement systems with the emphasis on the parity space is provided in [4]. However, it is imperative to acknowledge the inherent limitations of these approaches, particularly in detecting inaccuracies within in-range sensor values. Furthermore, the inclusion of redundant sensors introduces financial considerations, thereby warranting careful evaluation of the potential scenario where both sensors may concurrently exhibit inaccurate readings.

Model-based sensor diagnostics has emerged as a promising approach due to its ability to utilize the manipulator's dynamic model to detect and isolate sensor faults accurately. This approach relies on comparing the predicted sensor readings obtained from the model with the actual sensor measurements. Any discrepancies between the predicted and actual values indicate a potential sensor fault. Numerous studies have delved into the design of model based health monitoring and diagnostic systems. [5] proposed a method for isolating sensor bias faults in nonlinear systems based on adaptive thresholds. The use of analytical redundancy for detecting sensor failures in nuclear plants was investigated in [6]. In analytical redundancy, with the help of an assumed model of the physical system, the signals from a set of sensors are processed to reproduce the signals from all system sensors. [7] presented a robust model-based fault detection technique using adaptive robust observers. The existing general observer-based fault detection and isolation methods are dependent on partially known fault types. A fault detection and estimation scheme using the actual controller output and model-based compensation was developed in [8]. [9] used an adaptive extended Kalman filter to estimate the system states and the estimated outputs are compared with the measured signals to generate state residuals. In [10], [11], a bank of observers was used to isolate the sensor faults of the control system. These approaches are based on the controller design and may not be applicable everywhere.

Various other methods exist for diagnosing sensor faults in robot applications that exhibit the importance of sensor diagnostics in robotics. [12] presented a dynamic fault tolerance framework for remote robots, while [13] presented a sensor-driven, fault-tolerant control strategy for a maintenance robot, and [14] investigated the kinematic design of fault-tolerant manipulators.

These previous studies demonstrate the extensive research conducted on model-based sensor diagnostics for robotic manipulators, highlighting the importance of this field in ensuring the reliability, safety, and performance of robotic systems. This paper presents a methodology for systematically monitoring robot sensor readings corresponding to various measured quantities, ensuring mutual consistency through the imposition of one or more constraints. These constraints are derived from a model representing the physical system under observation, with each constraint encapsulating a relationship between two or more physical quantities. Consequently, the measurements of these physical quantities must adhere to these constraints. This approach eliminates the need for sensor redundancy, mitigates false negatives arising from multiple interdependent sensor readings, and facilitates the detection of erroneous sensor readings. Additionally, this methodology allows for the identification of faults in the robotic manipulator, which may result in constraint violations even in the presence of properly functioning sensors.

## II. DESCRIPTION

The model based sensor diagnostics methodology uses model constraints, which typically include model parameters that characterize the inherent properties of the various system components. System behavior is expressed in the form of mathematical equations (algebraic or differential). During robot operation, when a constraint is violated beyond an associated specified threshold, a sensor inconsistency is registered, indicating that at least one of the sensor readings in the constraint is erroneous. This suffices to initiate a fail-safe procedure. By adjusting the violation thresholds, error detection can be calibrated to different sensitivities, ranging from detecting only major errors to identifying more nuanced inconsistencies.

To monitor a complex system, e.g. robot manipulator, numerous constraints are utilized capturing fundamentally different aspects of the system. The system specifies a series elastic actuator (SEA) [15] for rotary joints on a robotic manipulator. A series elastic actuator for a rotary joint includes a motor (typically equipped with a position sensor or encoder), a gearbox, a series elastic element through which the load is transmitted (e.g. a torsional spring connected to the gearbox in series), a spring deflection sensor (SDS) measuring the angle that the spring deflects, and an output position sensor such as a magnetic angle encoder (MAE) on the output messing the output angle of the joint. The actuator is controlled by a conventional algorithm that sets the electrical current driving the motor. As shown in Fig. 1, are the model of the motor, gearbox, series spring through which the Load is transmitted. The load at a particular robotic joint includes all robot components at the end of the joint as well as any object the robot carries with the robotic manipulator.

The motor is typically equipped with a position sensor, such as a set of Hall-effect sensors, which measures the angular motor position, $\theta_M$. The motor also includes an Ampere-meter measuring the electrical drive current, $i_M$. The angular position from the gearbox output is not measured directly, and can be represented as $\theta_G$. A spring deflection sensor (SDS) with spring constant, $K_{SEA}$, measures the angle that the spring deflects, from which the spring torque, $\tau_{SEA}$, can be inferred. An output position sensor, such as a magnetic angle encoder (MAE) on the output, measures the output angle of the joint, i.e., the angular position, $\theta_L$ of the load.

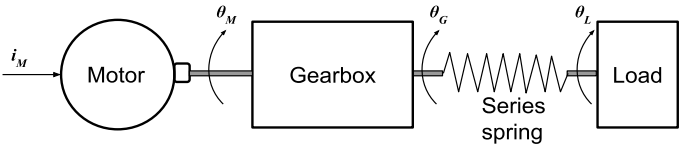

Fig. 1. Conceptual schematics of a series elastic actuator.

### III. MATHEMATICAL MODEL

For a robotic manipulator of various configurations, the constraints are derived from a physical model of the robot joint. The analytical model with mathematical equations typically include model parameters characterizing the physical properties and system behavior of the components. Alternatively, a computational model can be developed by optimizing the observed experimental data to the model predictions. The model can capture different aspects of the system behavior such as electrical, mechanical, electromechanical and/or thermal relationships between system components. The next section will discuss three such constraints imposed on a series elastic actuator of a robotic joint.
1. Torsional spring constraint
2. Joint dynamics constraint
3. Electrical motor constraint

These constraints capture the measured quantities of a robotic joint and their interdependence.

### A. Torsional Spring Constraint

The torsional spring constraint relates the load position, $\theta_L$ as measured by the MAE, the motor position ($\theta_m$) as measured by the motor Hall sensors, and the torque, $\tau_{SEA}$ as measured indirectly by the SDS. The model parameters of the torsional spring constraint include the gear ratio, $G_r$, and the overall stiffness of the system, $K_{Eq}$. The system's stiffness, $K_{Eq}$ depends on the spring constant, $K_{SEA}$ as well as the stiffness of the gear train, $K_{Gear}$; in a series arrangement, the inverses of the two stiffnesses simply add. These system parameters are known for the SEA, either as inferred from its design or as determined via characterization experiments.

The deflection of the spring, i.e., the relative deflection of its endpoints (that is, the load angle, $\theta_L$ on one end and the gearbox output angle, $\theta_G$ on the other end), is related to the torque, $\tau_{SEA}$ by the stiffness of the spring, $K_{SEA}$ as follows (where (t) indicates time-dependent quantities):

$$\theta_L(t) - \theta_G(t) = \frac{\tau_{SEA}(t)}{K_{SEA}} \quad (1)$$

Similarly, the deflection between the motor position ($\theta_m$) (adjusted by the gear ratio ($G_r$)) and the output position ($\theta_o$) is due to the deflection in the gear train resulting from the elasticity of the gear train. Because the gear train and spring are in series, they share the same torque $\tau$SEA, which is given by:

$$\theta_G(t) - \frac{\theta_M(t)}{G_r} = \frac{\tau_{SEA}(t)}{K_{Gear}} \quad (2)$$

Equations (1) and (2) yields,

$$\theta_L(t) - \frac{\theta_M(t)}{G_r} = \frac{\tau_{SEA}(t)}{K_{SEA}} + \frac{\tau_{SEA}(t)}{K_{Gear}} = \frac{\tau_{SEA}(t)}{K_{Eq}} \quad (3)$$

The above equation assumes no backlash from the gearbox.

### B. Joint Dynamics Constraint

The torque generated by the motor as sensed by the SDS, $\tau_{SEA}$, should be constrained by the current driven into the motor, $i_M$, the output position of the load, $\theta_L$, and the approximate second-order dynamics of the joint. The model parameters of this relationship include the inertia, $J$ and effective viscous damping, $B$ of the gearbox, the stiffness of the series spring,

$K_{SEA}$, as well as the motor torque constant, $K_T$ and the gear ratio, $G_r$.

When viewed from the output side of the gearbox, the gear train is subjected to damping torques captured by the damping factor $B$, torques due to the elasticity of the series spring, $K_{SEA}$ at the output, and the torques applied by the motor, $\tau_M$. The balance of these forces results in the following equation of motion:

$$\tau_M(t) = J\ddot{\theta}_G(t) + B\dot{\theta}_G(t) + K_{SEA}(\theta_G(t) - \theta_L(t))$$
$$= K_T G_r i_M(t) \quad (4)$$

The torque applied by the motor is primarily determined by the motor current, $i_M$, its torque constant, $K_T$ and the gear ratio, $G_r$. Since this equation of motion pertains to the output inertia of the gearbox, the motor torque is multiplied by the gear ratio to get an equivalent torque at the output.

The torque that is applied to the load, $\tau_L$ is equal to the torque in the series elastic element, $\tau_{SEA}$, and is given by:

$$\tau_L(t) = K_{SEA}(\theta_G(t) - \theta_L(t)) \quad (5)$$

The transfer function for the joint dynamics constraint can be derived from the above two equations by a Laplace transform, which converts the equations from time-domain differential equations to frequency-domain algebraic equations. The transformed equations are given by:

$$K_T G_r I_M(s)$$
$$= Js^2 \Theta_G(s) + Bs\Theta_G(s) + K_{SEA}(\Theta_G(s) - \Theta_L(s)) \quad (6)$$

and

$$\frac{T_L(s)}{K_{SEA}} + \Theta_L(s) = \Theta_G(s) \quad (7)$$

Substituting $\Theta_G(s)$ from (7) into (6), and manipulating, yields the transfer function of the system:

$$T_L(s) = \frac{K_T G_r K_{SEA}}{Js^2 + Bs + K_{SEA}} I(s) - \frac{K_{SEA}(Js^2 + Bs)}{Js^2 + Bs + K_{SEA}} \Theta_L(s) \quad (8)$$

where (s) denotes the Laplace differential operator. The first part of the transfer function characterizes the forward path dynamics of the system which is due to current control, and the second part characterizes the back-impedance of the system that is the torques experienced by the load due to load motion. Since the motor current, the load position and the torque ($\tau_L = \tau_{SEA}$) are continuously measured, the relationship between those quantities can be computed and verified in real-time.

## C. Electrical Motor Constraint

This constraint relates the voltage applied to the motor, $V_m$, the current measured in the motor, $i_M$, and the back electromotive force (EMF) as defined by the motor's velocity constant, $K_e$ and angular velocity, $\dot{\theta}_M$ based on the voltage balance at the motor.

A standard model for the motor is an RL-circuit, where the voltage $V_m$ applied to the motor by the driving circuit equals the sum of the resistive load characterized by resistance $R$, the inductive load characterized by inductivity $L$, and the back EMF $V_{EMF}$:

$$V_M(t) = i_M(t)R + L\frac{di_m(t)}{dt} + V_{EMF}(t) \quad (9)$$

The back EMF $V_{EMF}$ is directly related to the angular motor velocity $\dot{\theta}_M$ by the motor velocity constant $K_e$, resulting in:

$$V_M(t) = i_M(t)R + L\frac{di_m(t)}{dt} + K_e \dot{\theta}_M(t) \quad (10)$$

Further, for motors used in a robot SEA, the inductance of the motors is typically negligibly small, allowing the constraint to be simplified to:

$$V_M(t) = i_M(t)R + K_e \dot{\theta}_M(t) \quad (11)$$

This adequately captures the electrical motor constraint since the motor voltage, motor current and motor angular velocity are all measured.

## IV. DIAGNOSTIC METHODOLOGY

In an ideal scenario, with a perfect model and accurate sensor readings, these constraints should be precisely satisfied during robot operation. In practice, though, model imperfections and noise in sensor readings generally cause deviations from the constraints even in the absence of any sensor failure or other issues. Furthermore, sensor noise will generally result in frequent instantaneous violations of the constraints. To separate such transient noise-based violations from actor and permanent failure-based violation, the deviations from model constraints will be filtered with a second order low pass filter. The thresholds will be placed on the filtered errors. The integration time span of the filter will be tuned experimentally based on the noise levels of the sensors. The methodology is explained through the flowchart shown in Fig 2.

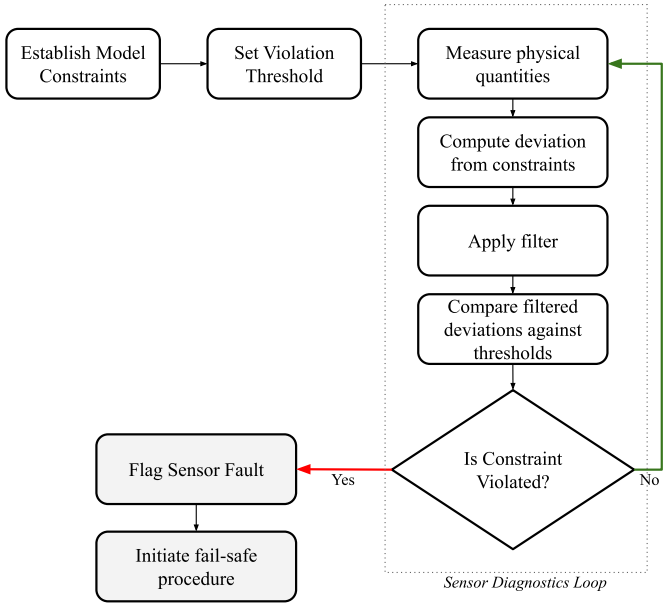

Fig. 2. Sensor Diagnostics Methodology

The sensor fault diagnostics is flagged when:
1. The torsional spring constraint is violated by

$$\phi_{MA}\left\{\left|K_{Eq}\left(\theta_L(t) - \frac{\theta_M(t)}{G_r}\right) - \tau_{SEA}(t)\right|\right\} \leq \varepsilon_{Torsional} \quad (12)$$

2. The joint dynamics constraint is violated by

$$\phi_{MA}\left\{\left|T_L(s) - \frac{K_T G_r K_{SEA}}{Js^2+Bs+K_{SEA}}I(s) + \frac{K_{SEA}(Js^2+Bs)}{Js^2+Bs+K_{SEA}}\Theta_L(s)\right|\right\}$$
$$\leq \varepsilon_{Dynamics} \quad (13)$$

3. The electrical motor constraint is violated by

$$\phi_{MA}\left\{\left|V_M(t) - i_M(t)R - K_e\dot{\theta}_M(t)\right|\right\} \leq \varepsilon_{Electrical} \quad (14)$$

In the (12), (13) and (14), $\phi_{MA}\{.\}$ indicates the filtered values of the quantities in braces. The violation thresholds $\varepsilon_{Torsional}$, $\varepsilon_{Dynamics}$ and $\varepsilon_{Electrical}$ are tuned based on empirical data. When the sensor values are smoothed and precise, and the models better capture the fidelity of the physical system, a lower constraint violation threshold can be applied.

## V. SIMULATION RESULTS

Consider a series elastic rotary joint comprising a motor, a gearbox, and a torsional spring connected in series between the load and the gearbox. The model is represented by (1). It is essential to highlight that the simulated model differs from the real physical model, and modeling uncertainty has been intentionally introduced into the simulation.

The real physical model is described by a higher-fidelity equation for the series elastic actuator, as given by:

$$\tau_{SEA} = K_1\left(\theta_L(t) - \frac{\theta_M(t)}{G_r}\right) + K_2\left(\theta_L(t) - \frac{\theta_M(t)}{G_1}\right)^2 \quad (15)$$

The simulations are executed with specific parameters: $K_1 = 100\ Nm/deg$, $K_2 = 0.02\ Nm/deg^2$, $G_1 = 105.05:1$, $K_{Eq} = 80$, and $G_r = 105$. The sensor readings from position sensors and torque sensors incorporate band-limited white noise. A low-pass filter with a 5 Hz cutoff frequency has been employed to smooth the noise from the residual of the torsional spring constraint equation. Consequently, the threshold for torsional spring faults is set at $\varepsilon_{Torsional} = 12$.

Two types of sensor faults have been considered which may occur in $\tau_{SEA}(t)$: bias in torque sensor reading, stuck sensor reading. Fig. 1 depicts the nominal behavior of the sensors. Fig. 2 and Fig. 3 illustrate cases of sensor faults and their detection. The residual generated by the sensor diagnostics algorithm and the threshold are depicted in the plots. Figure 2 shows the simulation results when a sensor fault in $\tau_{SEA}(t)$ occurs at $T_0 = 5\ s$, while Figure 3 shows the results when a sensor fault occurs at $T_0 = 3.1\ s$.

It is evident that the residual of the torsional spring constraint exceeds the threshold, indicating the presence of a fault in a sensor for $\tau_{SEA}(t)$. Similar results have been obtained for faults in sensors from joint dynamics constraint and electrical motor constraint, but due to space limitations, they are not reported here.

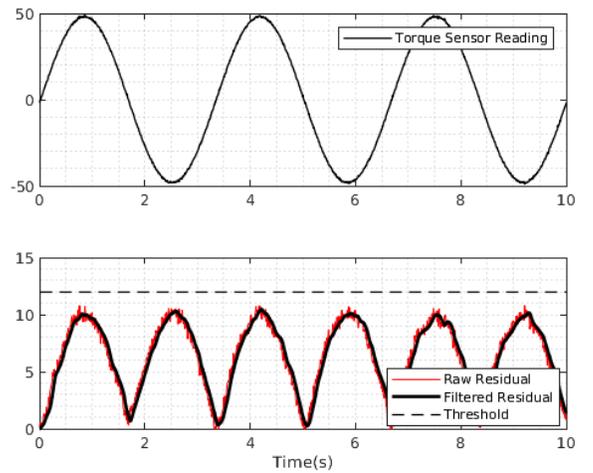

Fig. 3. Nominal sensor behavior

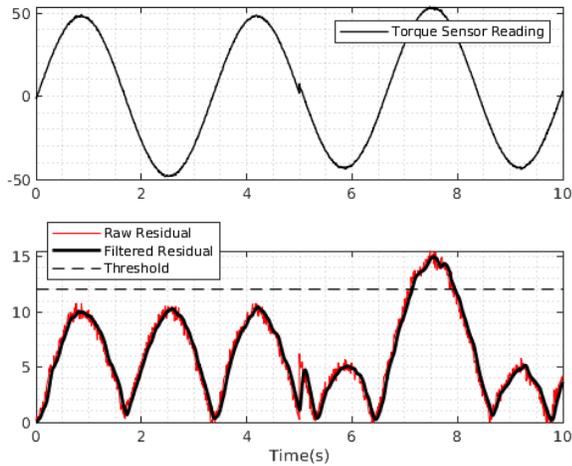

Fig. 4. Sensor fault due to bias

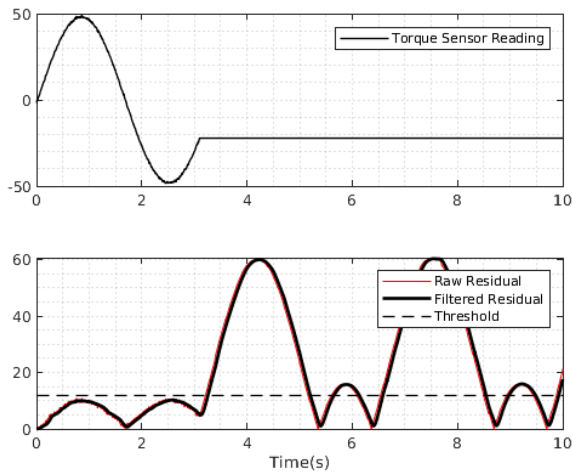

Fig. 5. Sensor fault due to sensor reading getting stuck

## VI. CONCLUSION

In conclusion, the application of model-based sensor diagnostics for robotic manipulators emerges as a crucial and effective tool for enhancing the safety and reliability of these intricate systems. This paper has introduced a novel methodology for crafting model-based constraints tailored for sensor diagnostics, characterized by analytical relationships spanning mechanical and electrical domains. A simulation example has been presented to illustrate the theoretical outcomes, underscoring the practical applicability of the proposed approach.

It is noteworthy that these constraints are contingent on specific modeling assumptions. Should there be a need to accommodate more accurate or higher fidelity models, the constraints can be readily adapted. For instance, in the case of series elastic actuators (SEAs), where the gearbox is initially modeled as an ideal element with a unique gear ratio, the introduction of higher fidelity models might consider resonant dynamics of the gear train, necessitating corresponding adjustments in the constraints.

In complex systems, the employment of multiple constraints becomes imperative to capture diverse aspects of the system or subsystem. The optimization of computational resources favors the encapsulation of relationships between monitored sensor readings within as few constraints as possible, resulting in a set of mutually independent constraints. However, it is essential to acknowledge that, in certain scenarios, the use of dependent constraints can be justified.

To broaden the scope of the proposed methodology, future work can explore the development of additional modeling constraints, extending beyond mechanical and electrical relationships to incorporate aspects such as thermal and electromechanical interactions. This expansion will contribute to a more comprehensive and robust framework for model-based sensor diagnostics in robotic manipulators, paving the way for continued advancements in their safety and reliability.